\documentclass[review]{elsarticle}

\usepackage{natbib}
\usepackage{amssymb}
\usepackage{longtable}
\usepackage{graphicx}
\usepackage{hyperref}
\usepackage{listings}
\newproof{pf}{Proof}

\usepackage{times}
\usepackage{url}
\usepackage{xcolor}
\usepackage{soul}
\usepackage[utf8]{inputenc}
\usepackage[small]{caption}
\usepackage{amssymb}
\usepackage{algorithm}
\usepackage[noend]{algpseudocode}
\usepackage{multirow}
\usepackage{subcaption}
\usepackage{tabularx}
\usepackage{booktabs}
\usepackage[flushleft]{threeparttable}

\newcommand{\alctr}{\mathcal{ALC}+\tip_{\bf R}}

\newcommand{\tip}{{\bf T}}
\newcommand{\alc}{\mathcal{ALC}}
\newcommand{\alct}{\mathcal{ALC}+\tip_{\bf R}}

\newcommand {\kk} {\mathcal{K}}
\newcommand{\RR}{{\mathcal{R}}}
\newcommand{\TT}{{\mathcal{T}}}
\newcommand{\AAA}{{\mathcal{A}}}
\newcommand{\pfl}{\tip^{\textsf{\tiny CL}}}

\newcommand{\II}{{\mathcal{I}}}
\newcommand{\emme}{{\mathcal{M}}}

\newtheorem{theorem}{Theorem}[section]

\newtheorem{definition}[theorem]{Definition}

\begin{document}

\begin{frontmatter}



\title{A Commonsense Reasoning Framework for Explanatory Emotion Attribution, Generation and Re-classification}


\author[UniTOaddress,ICARaddress]{Antonio Lieto}
\author[UniTOaddress]{Gian Luca Pozzato}
\author[UniTOaddress]{Stefano Zoia} 
\author[UniTOaddress]{Viviana Patti}
\author[UniTOaddress]{Rossana Damiano}

\address[UniTOaddress]{University of Turin, Department of Computer Science, Turin, Italy}
\address[ICARaddress]{ICAR-CNR, Palermo, Italy}

\begin{abstract}

We present DEGARI  (Dynamic  Emotion  Generator  And  ReclassIfier), an explainable system for emotion attribution and recommendation. This system relies on a recently introduced commonsense reasoning framework, the $\pfl$ logic, which is based on a human-like procedure for the automatic generation of novel concepts in a Description Logics knowledge base. Starting from an ontological formalization of emotions based on the Plutchik model, known as ArsEmotica, the system exploits the logic $\pfl$ to automatically generate novel commonsense semantic representations of compound emotions (e.g. Love as derived from the combination of Joy and Trust according to Plutchik). The generated emotions correspond to prototypes, i.e. commonsense representations of given concepts, and have been used to reclassify emotion-related contents in a variety of artistic domains, ranging from art datasets to the editorial contents available in RaiPlay, the online platform of RAI  Radiotelevisione Italiana (the Italian public broadcasting company). 
We show how the reported results (evaluated in the light of the obtained reclassifications, the user ratings assigned to such reclassifications, and their explainability) are encouraging, and pave the way to many  further research directions. 

\end{abstract}

\end{frontmatter}

\section{Introduction and Background}




Emotions have been acknowledged as a key part of the aesthetic experience through all ages and cultures, 
as witnessed by terms such as ``sublime" \cite{russell1964sublime} and ``pathos" \cite{ross1959ars}, associated with the experience of art since the ancient times. 
The advent of computational tools and methods for investigating the way we respond emotionally to objects and situations has paved the way to a deeper understanding of the intricate relationship between emotions and art. For example, \citeauthor{van2016implicit} \cite{van2016implicit} have studied how art affects emotional regulation by measuring the brain response through EEG: their research shows that, in comparison with photographs depicting real events, artworks determine stronger electro-physiological responses; in parallel, \citeauthor{leder2014makes} \cite{leder2014makes} argue that the emotional response to art -- measured through facial muscle movements -- is attenuated in art critics, and stronger in  non-experts, thus showing the universality and spontaneity of this response.

The association between art and emotions is even stronger when the artistic expression is conveyed by non-textual media, as in music and, at least partly, in movies. For example, music has proven to be an effective tool for emotion regulation: as demonstrated by \citeauthor{thoma2012emotion} \cite{thoma2012emotion}, music can induce specific emotional states in everyday situations, an effect which is sought for by the users and can be exploited to create effective affective recommender systems \cite{andjelkovic2019moodplay}.
Finally, emotional engagement is of primary importance in narrative media, such as film and television, as extensively investigated by a line of research which draws from both film studies and emotion theories  \cite{smith2003film,tan2013emotion}.

\medskip

As a consequence of the multi-faceted, complex role played by emotions in the experience of art and media, the investigation of this phenomenon with computational tools has relied on a variety of models and methodologies, ranging from dimensional models, better suited to investigate physiological, continuous correlate of emotions \cite{russell1980circumplex,watson1988development,mehrabian1996pleasure}, to categorical models, which lend themselves to inspecting the conscious level of emotional experience \cite{plutchik1980general,ekman1999basic,banziger2010introducing}. Dimensional models  typically measure the emotional engagement along the arousal and hedonic axes, and are useful to study how the emotional response evolves over time.
For example, \citeauthor{lopes2017modelling} \cite{lopes2017modelling} rely on crowdsourced annotations of tension, arousal and variance in audio pieces to realize sound-based affective interaction in games. 
Categorical models are useful to collect the audience experience as discrete emotional labels, and are easily mapped to textual descriptions of emotions across languages. %
As exemplified by \citeauthor{mohammad-kiritchenko-2018-wikiart}  \cite{mohammad-kiritchenko-2018-wikiart}, discrete emotional labels, merged from different categorical models \cite{plutchik1980general,noy2013art}, can shed light on the reception of art, letting correlations emerge between attributed emotions, liking and subjects.

In many cases, the emotional response  of the audience is conveyed through language, non only in textual media, but also in relation to art and other media: consider, for example, tags and social media comments concerning artworks and exhibitions.
Automatically detecting affective states and emotions from text has gained considerable attention over recent years, leading to the development of several resources - such as annotated corpora, ontologies and lexicons within the Computational Linguistics community \cite{JurafskyM09,NISSIM201731,CambriaReview20}, {also in response to the spread of supervised
methods requiring a large amount of training data.
Recently, deep neural network models have attracted increasing attention, and are being applied also to tasks related to the detection of affective states, obtaining promising results. Several neural architectures have been developed for a variety of tasks ranging from 
emotion detection \cite{CHATTERJEE2019309}, to dimensional sentiment analysis \cite{wang-etal-2016-dimensional} and the most common sentiment polarity detection task, and have been evaluated against datasets on different types of social media texts, including long reviews, or, most of the time, short microblog messages, such as tweets. Interestingly, attention-based deep models turned out to be particularly effective, achieving state-of-the-art results on both long review and short tweet polarity classification. This is the case of the  attention-based bidirectional CNN-RNN deep model for
sentiment analysis in \cite{ABCDM-BASIRI2021279}, showing that applying an attention layer to the outputs of the LSTM and GRU branches of the network makes the semantic
representations more informative.}

Affective information expressed in texts is multi-faceted, and the wide variety of affective linguistic resources developed in the last years, mainly for English, but also for other languages, basically reflects such richness. When we speak about {\em affective states} in the context of natural language communication, we mean to refer to several aspects, which vary in their degree of stability, such as: emotion, sentiment, personality, mood, attitudes or interpersonal stance \cite{mohammad2020survey}. 
Given the wide variety of affective states, in recent years, research has focused on a finer-grained investigation of the role of emotions, as well as on the importance of other affect dimensions such as sentiment and emotion intensity \cite{mohammad2017word,cambria-ieee-intense20} or activation.
{On this line, recent efforts on predicting the degree of intensity for emotion and sentiment in different domains  led to interesting experiments devoted to effectively combining deep learning and feature driven traditional models via an ensemble framework approach  \cite{cambria-ieee-intense20}}.

Depending on the specific research goals addressed, one could be interested in issuing a discrete label describing the affective state expressed (frustration, anger, joy, etc.) to address different contexts of interaction and tasks. 
Both basic emotion theories, in the Plutchik-Ekman \cite{ekman1999basic} tradition, and dimensional models of emotions, have provided a precious theoretical grounding for the development of lexical resources \cite{StrappaRada07,Mohammad13,mohammad-2018-vad,cambria2020senticnet,strapparava-valitutti-2004-wordnet} and computational models for emotion extraction. However, there is a general tendency to move towards richer, finer-grained models, possibly including complex emotions, especially in the context of data-driven and task-driven approaches, where restricting the automatic detection to a small set of basic emotions would fall short to achieve the objective. This is also our perspective. 

From a computational point of view, 
the choice of the model of affect to be used in order to give psychological grounding to the resource or the corpus to be developed is driven from, and highly intertwined with, the specific sentiment analysis task to be addressed, which, in turn, usually depends on the application domain to be tackled and on the final purpose of mining affective contents in texts.
In this sense, 
evaluating the emotional responses of an audience in front of an artwork, with the purpose of monitoring the emotional impact of a cultural heritage exhibition on visitors \cite{BertolaP16}, 
is different from monitoring political sentiment or mining the levels of anger in comments threads of software developers \cite{Novielli:2017}.
There are still few works and resources specifically developed to address emotion detection in the art and media domain. These include the work by \citeauthor{mohammad-kiritchenko-2018-wikiart}  \cite{mohammad-kiritchenko-2018-wikiart}, where the authors describe the WikiArt Emotions Dataset, which includes emotion annotations for thousands of pieces of art from the WikiArt.org collection, and the work by Patti et al. \cite{BertolaP16,patti2015arsemotica} where the ArsEmotica framework\footnote{Available at \url{http://130.192.212.225/fuseki/ArsEmotica-core}} is proposed, which relies on the combined use of NLP affect resources and an ontology of emotions to enable an emotion-driven exploration of online art collections.


The diversity of computational models implied by the analysis of the emotional response to art and media, and of the applications that exploit this response to improve the user experience -- from learning to entertainment -- witnesses the complexity of the underlying processes (including, but not limited to, aesthetic, self-regulatory, social and cultural processes). This diversity, however, can be an obstacle to the development of models which work across domains and formats, preventing techniques from being transferred across similar tasks (e.g., emotion annotation and affective recommendation). In particular, the differences in emotion annotation between datasets can endanger the development and cross-validation of new techniques for analysing and exploiting emotions in art and media. In this sense, techniques for merging and extending emotional categories can be useful to overcome these limitations. 
A notable example of such a comprehensive system is SenticNet \cite{cambria2020senticnet}, which relies on the Hourglass model \cite{cambria2012hourglass}. 
The Hourglass model, recently revised and extended \cite{susanto2020hourglass}, is inspired by Plutchik’s model  of emotions  \cite{plutchik2001nature}. Such model, formalized in the ArsEmotica ontology and described in detail in Section \ref{arsemotica}, can be represented as a wheel of emotions, formed by: basic or primary emotions; opposite emotions; similarity between the emotions; compound emotions (or complex emotions) generated by the primary ones.
Similarly to the SenticNet framework, our system also relies on  the Plutchik model. The choice of this model is based on the fact that it provides a recipe for the generation of compound emotions that is compliant with the commonsense reasoning framework of the $\pfl$ logic. %
As such, we exploited the reasoning mechanisms of $\pfl$ to generate the compound emotions according to Plutchik's theory.

In this paper, we illustrate and validate this approach by means of the DEGARI system (Dynamic Emotion Generator And ReclassIfier) for emotion attribution and recommendation. 
In particular, we exploit the compound concepts generated by the system to automatically reclassify items in three datasets in the artistic and media domains. As a result of this reclassification process, an \emph{emotional enrichment} is obtained and new emotional labels are associated with the items in the original datasets.
To the best of our knowledge, DEGARI is the first emotion-oriented system employing a white box approach to emotion classification based on the human-like conceptual combination framework proposed in the $\pfl$ logic.  DEGARI is available at \url{http://di.unito.it/DEGARI}.

The key contributions provided in this work are the following:

\begin{itemize}
    \item an entirely explainable AI system for automatic emotion re-classification and recommendation  based on a well founded emotion theory (Plutchik model) and on a probabilistic logic framework modelling human-like for concept combination (i.e. the $\pfl$ logic);
    \item the ontology of the Plutchik emotion model (made available as a free resource at \url{http://130.192.212.225/fuseki/ArsEmotica-core} and queryable via a SPARQL endpoint at \url{http://130.192.212.225/fuseki/dataset.html?tab=query&ds=/ArsEmotica-core}); 
    \item an empirical and replicable validation (i.e. data and software are publicly available) of the proposed framework on three different datasets covering diverse artistic domains: multimedia, paintings and miscellaneous artistic items (poems,  videos,  pictures and music).
\end{itemize}
\normalcolor





The paper is organized as follows: after a brief overview of the rationale adopted by our commonsense reasoning framework (Section~\ref{rationale}), we present in Section~\ref{logic} - for the sake of self-containedness - a more detailed description of the $\pfl$ logic (by referring to \cite{lieto2018description} for a complete explanation). In Section \ref{arsemotica} we present the ontological model ArsEmotica (enriched with an emotional lexicon) formalizing Plutchik's theory of emotions and used as a standard representation to leverage the reasoning capabilities of the  $\pfl$ within the system DEGARI. Sections \ref{degari1} and \ref{degari2} present the DEGARI system that, starting from the basic emotions represented in ArsEmotica (and according to  Plutchik's theory), generates compound emotions and uses these novel emotional categories for artistic content reclassification.  
Section~\ref{explsection} discusses how DEGARI can be considered as an explainable AI system.
Finally, Section~\ref{evaluation} shows the outcome of the automatic and explainable reclassification obtained with DEGARI (exploited in different settings and with different affective lexicons) and the results of a user study on $44$ people showing the feasibility of using the obtained reclassifications as recommended labels. Section~\ref{final} ends the paper.


\section{Commonsense Concept Invention via Dynamic Knowledge Combination}\label{rationale}

The overall rationale assumed in the $\pfl$ reasoning framework is that the process of automatic generation of novel concepts within a knowledge base (also known as \emph{knowledge invention} process) can be obtained, as happens in humans \cite{lieto2018description,ismis2018}, by exploiting a process of commonsense conceptual combination. This generative phenomenon highlights some crucial aspects of the knowledge capabilities in human cognition. Such ability, in fact, is associated to creative thinking and problem solving. Still, however, it represents an open challenge in the field of Artificial Intelligence (AI) \cite{boden1998creativity}. 
Dealing with this problem, indeed, requires, from an AI  perspective, the harmonization of two conflicting requirements that are hardly accommodated in symbolic systems \cite{frixione2011representing}: the need for a syntactic and semantic compositionality (typical of logical systems) and the one concerning the exhibition of typicality effects.  
According to a well-known argument \cite{osherson1981adequacy}, in fact, prototypes (i.e. commonsense conceptual representations based on typical properties) are not compositional. The argument runs as follows: consider a concept like \emph{pet fish}. It results from the composition of the concept \emph{pet} and of the concept \emph{fish}. However, the prototype of \emph{pet fish} cannot result from the composition of the prototypes of a pet and a fish: e.g., a typical pet is furry and warm, a typical fish is grayish, but a typical pet fish is neither furry and warm nor grayish (typically, it is red). The \emph{pet fish} phenomenon is a paradigmatic example of the difficulty to address when building formalisms and systems trying to imitate this combinatorial human ability. 
In this paper, we exploit the recently introduced nonmonotonic extension of Description Logics $\pfl$ (typicality-based compositional logic, introduced in \cite{lieto2018description}), which is able to account for this type of human-like concept combination \footnote{Other works have already shown how such logic can be used to model complex cognitive phenomena \cite{lieto2018description}, goal-directed creative problem solving \cite{kcap, csr19beyond, ecai2} and to build intelligent applications 
for
computational creativity \cite{lietopozzato2019creativityia}. Alternative approaches to the problem of commonsense conceptual combination are discussed in \cite{eppe2018computational}, \cite{lewis2016hierarchical}, \cite{confalonieri2016}. The advantages of $\pfl$ with respect to such approaches are detailed in \cite{lieto2018description}.}. More specifically, we show how it can be used as a tool for the generation of novel compound emotions and, as a consequence, for the suggestion of novel emotion-related contents.

In $\pfl$, ``typical'' properties can be directly specified  by means of a
``typicality'' operator $\tip$ enriching  the underlying Description Logic (from now on, DL for short), and a TBox can contain inclusions of the form $\tip(C) \sqsubseteq D$ to represent that ``typical $C$s are also $Ds$''. As a difference with standard DLs, in the logic $\pfl$  one can consistently express exceptions and reason about defeasible inheritance as well. Typicality inclusions are also equipped by a real number $p \in (0.5,1]$ representing the probability/degree of belief in such a typical property: this allows us to define a semantics inspired to the DISPONTE semantics \cite{riguzzi} characterizing probabilistic extensions of DLs, which in turn is used in order to describe different \emph{scenarios} where only some typicality properties are considered. Given a KB containing the description of two concepts $C_H$  and $C_M$ occurring in it, we then consider only {\em some}  scenarios in order to define a revised knowledge base, enriched by typical properties of the combined concept $C \sqsubseteq C_H \sqcap C_M$ by also implementing a heuristics coming from the cognitive semantics.

By relying on $\pfl$, this work introduces the system DEGARI which, first, automatically builds prototypes of existing \emph{compound} emotions  by extracting information about \emph{concepts} or \emph{properties} relying on the ArsEmotica ontology enriched with the NRC Emotion Intensity Lexicon \cite{mohammad2017word} (this lexicon associates, in descending order of frequency, words to emotional concepts). In this setting, words with the highest frequencies of association to emotional concepts have been used as typical features of the basic emotions in the Plutchik model.  Such prototypes of basic emotions have been formalized by means of a $\pfl$ knowledge base, whose TBox contains both \emph{rigid} inclusions of the form $$\mathit{BasicEmotion} \sqsubseteq \mathit{Concept},$$ in order to express essential desiderata but also constraints, as an example $\mathit{Joy} \sqsubseteq \mathit{PositiveEmotion}$ as well as \emph{prototypical} properties of the form $$p \ :: \ \tip(\mathit{BasicEmotion}) \sqsubseteq \mathit{TypicalConcept},$$  representing typical concepts of a given emotion, where $p$ is a real number in the range $(0.5,1]$, expressing the frequency of such a concept in items belonging to that emotion: for instance, $0.72 \ :: \ \tip(\mathit{Surprise}) \sqsubseteq \mathit{Delight}$ is used to express that the typical feature of being surprised contains/refers to the emotional concept {\em Delight} with a frequency/probability/degree of belief of the $72\%$.

Given the ArsEmotica knowledge base (see Section~\ref{arsemotica}) equipped with the prototypical descriptions of basic emotions, DEGARI exploits the reasoning capabilities of the logic $\pfl$ in order to generate new \emph{derived} emotions as the result of the creative combination of two (or even more)  basic or derived ones.
DEGARI also reclassifies the artistic and multimedia contents taking the new, derived emotions into account. Intuitively, an item of the tested dataset belongs to the new generated emotion if its metadata (name, description, title) contain all the rigid properties as well as at least the $30\%$ of the typical properties of such a derived emotion. 
In this respect, DEGARI can be seen as a ``white box'' recommender system, able to suggest to its users artistic contents belonging to new emotions by providing an explanation of such a recommendation.

We have tested DEGARI by performing different kinds of evaluation that are reported and discussed in Section \ref{evaluation}. In the following, 
in order to make the paper self-contained
we recall in more detail the main features of the above described $\pfl$  logic. 


\section{THE DESCRIPTION LOGIC $\pfl$ FOR CONCEPT COMBINATION}
\label{logic}
The logic $\pfl$ \cite{lieto2018description} used by the system DEGARI as the basis for the generation of new compound emotions combines three main ingredients. 
The first one relies on the DL of typicality $\alct$ introduced in \cite{AIJ2014}, which allows to describe the \emph{protoype} of a concept.
In this logic, ``typical'' properties can be directly specified  by means of a
``typicality'' operator $\tip$ enriching  the underlying DL, and a TBox can contain inclusions of the form $\tip(C) \sqsubseteq D$ to represent that ``typical $C$s are also $Ds$''. As a difference with standard DLs, in the logic $\alct$  one can consistently express exceptions and reason about defeasible inheritance as well. 
For instance, a knowledge base can consistently express that ``normally, athletes are fit'', whereas ``sumo wrestlers usually are not fit'' by  $\tip (\mathit{Athlete}) \sqsubseteq \mathit{Fit}$ and $\tip (\mathit{SumoWrestler}) \sqsubseteq  \lnot \mathit{Fit}$, given that $\mathit{SumoWrestler} \sqsubseteq \mathit{Athlete}$. 
The semantics of the $\tip$ operator is  characterized by the properties of \emph{rational logic} \cite{whatdoes}, recognized as the core properties of nonmonotonic reasoning. $\alct$ is characterized by a minimal model semantics corresponding to an extension to DLs of a notion of \emph{rational closure} as defined in \cite{whatdoes} for propositional logic: the
idea is to adopt a preference relation over $\alct$ models, where intuitively a model is preferred to another one if it contains less exceptional elements, as well as a notion of \emph{minimal entailment} restricted to models that are minimal with respect to such preference relation.
As a consequence, $\tip$ inherits well-established properties like \emph{specificity} and \emph{irrelevance}: in the example, the logic $\alct$ allows us to infer $\tip(\mathit{Athlete} \sqcap \mathit{Bald}) \sqsubseteq \mathit{Fit}$ (being bald is irrelevant with respect to being  fit) and, if one knows that Hiroyuki is a  typical sumo wrestler, to infer that he is not fit, giving preference to the most specific information.

A second ingredient consists of a distributed semantics similar to the one of probabilistic DLs known as DISPONTE \cite{disponteijcai}, which allows labeling inclusions $\tip(C) \sqsubseteq D$ with a real number between 0.5 and 1, which represents its degree of belief/probability, under the assumption that each axiom is independent from each others. Degrees of belief in typicality inclusions allow defining a probability distribution over \emph{scenarios}: roughly speaking, a scenario is obtained by choosing, for each typicality inclusion, whether it is considered as true or false.
In a slight extension of the above example, we could have the need to represent that both the typicality inclusions about athletes and sumo wrestlers have a degree of belief of  $80\%$, whereas we also believe that athletes  are usually young with a higher degree of $95\%$, with the following KB:
\begin{quote}
(1) \ $\mathit{SumoWrestler} \sqsubseteq \mathit{Athlete}$\\
(2) \ $0.8 \ :: \ \tip (\mathit{Athlete}) \sqsubseteq \mathit{Fit}$\\
(3) \ $0.8 \ :: \ \tip (\mathit{SumoWrestler}) \sqsubseteq  \lnot \mathit{Fit}$\\
(4) \ $0.95 \ :: \ \tip (\mathit{Athlete}) \sqsubseteq \mathit{YoungPerson}$
\end{quote}
In this case, we consider eight different scenarios,
representing all possible combinations of typicality inclusion: as an example, $\{ ((2),1), ((3),0), ((4),1) \}$ represents the scenario in which (2) and (4) hold, whereas (3) does not. Obviously, (1) holds in every scenario, since it represents a rigid property, not admitting exceptions. We  equip each scenario with a  probability depending on those of the involved  inclusions: the scenario of the example has probability $0.8 \times 0.95$ (since 2 and 4 are involved) $\times (1-0.8)$ (since 3 is not involved) $ = 0.152 = 15.2\%$. Such probabilities are then taken into account in order to choose the most adequate scenario describing the prototype of the combined concept.

As a third element of the proposed formalization is a method inspired by cognitive semantics \cite{hampton1987inheritance} for the identification of a dominance effect between the concepts to be combined: for every combination, we distinguish a  HEAD, representing the stronger element of the combination, and a MODIFIER.  The basic idea is: given a KB and two concepts $C_H$ (HEAD) and $C_M$ (MODIFIER) occurring in it, we consider only {\em some} scenarios in order to define a revised knowledge base, enriched by typical properties of the combined concept $C \sqsubseteq C_H \sqcap C_M$.

Let us now present the logic $\pfl$ more precisely.
The language of $\pfl$ extends the basic DL $\mathcal{ALC}$ by \emph{typicality inclusions} of the form $\tip(C) \sqsubseteq D$ equipped by a real number $p \in (0.5,1]$ -- observe that the extreme $0.5$ is not included -- representing its degree of belief, whose  meaning is that ``we believe with degree/probability $p$ that, normally, $C$s are also $D$s'' \footnote{The reason why we only allow typicality inclusions equipped with probabilities $p > 0.5$ is due to our effort of integrating two different semantics: typicality based logic and DISPONTE. In particular, as detailed in \cite{lieto2018description} this choice seems to be the only one compliant with both formalisms. On the contrary, it would be misleading to also allow low degrees of belief for typicality inclusions, since typical knowledge is known to come with a low degree of uncertainty.}

\begin{definition}[Language of $\pfl$]
We consider an alphabet of concept names $\mathtt{C}$, of role names
$\mathtt{R}$, and of individual constants $\mathtt{O}$.
Given $A \in \mathtt{C}$ and $R \in \mathtt{R}$, we define:

\vspace{0.2cm}

 $ C, D:= A \mid \top \mid \bot \mid  \lnot  C \mid  C \sqcap  C \mid  C \sqcup  C \mid \forall R. C \mid \exists R. C$
 
\vspace{0.2cm}

\noindent   We define a knowledge base $\kk=\langle \RR, \TT, \AAA \rangle$ where:

\noindent $\bullet$ $\RR$ is a finite set of rigid properties of the form $C \sqsubseteq D$;
    
\noindent $\bullet$ $\TT$ is a finite set of typicality properties of the form $$p \ :: \ \tip(C) \sqsubseteq D$$ where $p \in (0.5,1] \subseteq \mathbb{R}$ is the degree of belief of the typicality inclusion;
 
 \noindent $\bullet$ $\AAA$ is the ABox, i.e. a finite set of formulas of the form either $C(a)$ or $R(a,b)$, where $a, b \in \mathtt{O}$ and $R \in \mathtt{R}$.
\end{definition}

A model $\emme$ in the logic $\pfl$ extends standard $\mathcal{ALC}$ models by a preference relation among domain elements as in the logic of typicality \cite{AIJ2014}. In this respect, $x < y$ means that $x$ is ``more normal'' than $y$, and that the typical members of a concept $C$ are the minimal elements of $C$ with respect to this relation\footnote{It could be possible to consider an alternative semantics whose models are equipped with multiple preference relations. However the approach based on a single preference relation in \cite{AIJ2014} ensures good computational properties (reasoning in the resulting nonmonotonic logic $\alct$ has the same complexity of the standard $\alc$), whereas adopting multiple preference relations could lead to higher complexities.}. An element $x
\in \Delta^\II$ is a {\em typical instance} of some concept $C$ if $x \in
C^\II$ and there is no $C$-element in $\Delta^\II$ {\em more normal} than
$x$. Formally:

\begin{definition}[Model of $\pfl$]
A model $\emme$ is any
structure $$\langle \Delta^\II, <, .^\II \rangle$$ where: 
\begin{itemize}
\item $\Delta^\II$ is a non empty set of items called the domain;
\item $<$ is an irreflexive, transitive, well-founded and modular (for all $x, y, z$ in $\Delta^\II$, if
$x < y$ then either $x < z$ or $z < y$) relation over
$\Delta^\II$;
\item $.^\II$ is the extension function that maps each atomic
concept $C$ to $C^\II \subseteq \Delta^\II$, and each role $R$
to  $R^\II \subseteq \Delta^\II \times \Delta^\II$, and is extended to complex concepts as follows:

\begin{itemize}
   \item $(\lnot C)^\II = \Delta^\II \ \backslash \ C^\II$
   \item $(C \sqcap D)^\II = C^\II \cap D^\II$
   \item $(C \sqcup D)^\II = C^\II \cup D^\II$
   \item $(\exists R.C)^\II = \{x \in \Delta^\II \mid \exists (x,y) \in R^\II \ \mbox{such that} \ y \in C^\II \}$
   \item $(\forall R.C)^\II = \{x \in \Delta^\II \mid \forall (x,y) \in R^\II \ \mbox{we have} \ y \in C^\II \}$
   \item $(\tip(C))^\II = Min_<(C^\II),$ where $Min_<(C^\II)=\{x \in C^\II \mid \nexists y \in C^\II \ \mbox{s.t.}  \ y<x\}$.
\end{itemize}
\end{itemize}
\end{definition}

\noindent A model $\emme$  can be equivalently defined by postulating the existence of
a function $k_{\emme}: \Delta^\II \longmapsto \mathbb{N}$, where $k_{\emme}$ assigns a finite rank to each domain element \cite{AIJ2014}: the rank of $x$ is the
length of the longest chain $x_0 < \dots < x$ from $x$
to a minimal $x_0$, i.e. such that there is no ${x'}$ such that  ${x'} < x_0$. The rank function $k_{\emme}$ and $<$ can be defined from each other  by letting $x < y$ if and only if $k_{\emme}(x) < k_{\emme}(y)$.


\begin{definition}[Model satisfying a knowledge base in $\pfl$]\label{semantica}
Let $\kk=\langle \RR, \TT, \AAA \rangle$
be a KB. Given a model $\emme=\langle \Delta^{\mathcal{I}}, <, .^{\mathcal{I}}\rangle$, we assume that $.^\II$ is extended  to assign a  domain element
$a^{\mathcal{I}}$ of $\Delta^\II$ to each individual constant $a$ of $\mathtt{O}$.
We say that:
\begin{itemize}
\item $\emme$  satisfies $\RR$ if, for all $C \sqsubseteq D \in \RR$, we have $C^\II \subseteq D^\II$;
\item $\emme$ satisfies $\TT$ if, for all $q \ :: \ \tip(C) \sqsubseteq D \in \TT$, we have that 
\footnote{It is worth noticing that here the degree $q$ does not play any role. Indeed, a typicality inclusion $\tip(C) \sqsubseteq D$ holds in a model only if it satisfies the semantic condition of the underlying DL of typicality, i.e. minimal (typical) elements of $C$ are elements of $D$. The degree of belief $q$ will have a crucial role in the application of the distributed semantics, allowing the definition of scenarios as well as the computation of their probabilities.}
$\tip(C)^\II \subseteq D^\II$, i.e. $Min_<(C^\II) \subseteq D^\II$; 
\item $\emme$ satisfies $\AAA$ if, for each assertion $F \in \AAA$, if $F = C(a)$ then $a^{\mathcal{I}} \in C^{\mathcal{I}}$, otherwise
          if $F = R(a,b)$ then $(a^{\mathcal{I}},b^{\mathcal{I}}) \in R^{\mathcal{I}}$.
\end{itemize}
\end{definition}

Even if the typicality operator $\tip$ itself  is nonmonotonic (i.e.
$\tip(C) \sqsubseteq E$ does not imply $\tip(C \sqcap D)
\sqsubseteq E$), what is inferred
from a KB can still be inferred from any KB' with KB $\subseteq$
KB', i.e. the resulting logic is monotonic. As already mentioned, in order to perform useful nonmonotonic inferences,   in
\cite{AIJ2014} the authors have strengthened  the above semantics by
restricting entailment to a class of minimal 
models. Intuitively, the idea is to
restrict entailment to models that \emph{minimize the atypical instances of a concept}. The resulting logic corresponds to a notion of \emph{rational closure} on top of $\alct$. Such a notion is a natural extension of the rational closure construction provided  in \cite{whatdoes} for the propositional logic.  This nonmonotonic semantics relies on minimal rational models  that minimize the \emph{rank  of domain elements}. Informally, given two models of KB, one in which a given domain element $x$ has rank 2 (because for instance
$z < y < x)$, and another in which it has rank 1 (because only
$y < x$), we prefer the latter,
as in this model the element $x$ is assumed to be ``more typical'' than in the former.
Query entailment is then restricted to minimal {\em canonical models}. The intuition is that a canonical model contains all the individuals that enjoy properties that are consistent with KB. This is needed when reasoning about the rank of the concepts: it is important to have them all represented.

 Given a KB $\kk=\langle \RR, \TT, \AAA \rangle$ and given two concepts $C_H$ and $C_M$ occurring in $\kk$, the logic $\pfl$ allows defining a prototype of the combined concept $C$ as the combination of the HEAD $C_H$ and the MODIFIER $C_M$, where  the typical properties of the form $\tip(C) \sqsubseteq D$ (or, equivalently, $\tip(C_H \sqcap C_M) \sqsubseteq D$) to be ascribed to the concept $C$ are obtained by considering blocks of scenarios with the same probability, in decreasing order starting from the highest one. We first discard all the inconsistent scenarios, then:
\begin{itemize}
   \item we discard those scenarios considered as \emph{trivial}, consistently inheriting  all the properties from the HEAD from the starting concepts to be combined. This choice is motivated by the challenges provided by task of commonsense conceptual combination itself: in order to generate plausible and creative compounds, it is necessary to maintain a level of surprise in the combination. Thus both scenarios inheriting all the properties of the two concepts and all the properties of the HEAD are discarded, since they prevent this surprise;
   \item among the remaining ones, we discard those inheriting properties from the MODIFIER which are in conflict with properties that could be consistently inherited from the HEAD;
   \item if the set of scenarios of the current block is empty, i.e. all the scenarios have been discarded either because trivial or because  the MODIFIER is preferred, we repeat the procedure by considering the block of scenarios  having the immediately lower probability.
\end{itemize}
\noindent Remaining scenarios are those selected by the logic $\pfl$.
 The ultimate output of our mechanism is a knowledge base in the logic $\pfl$ whose set of typicality properties is enriched by those of the compound concept $C$.
Given a scenario $w$ satisfying the above properties, we define the properties of  $C$ as the set of inclusions $p \ :: \ \tip(C) \sqsubseteq D$, for all $\tip(C) \sqsubseteq D$ that are entailed from $w$ in the logic $\pfl$. The probability $p$ is such that:
\begin{itemize}
 \item if $\tip(C_H) \sqsubseteq D$ is entailed from $w$, that is to say $D$ is a property inherited either from the HEAD (or from both the HEAD and the MODIFIER),  then $p$ corresponds to the degree of belief of such inclusion of the HEAD in the initial knowledge base, i.e. $p \:: \ \tip(C_H) \sqsubseteq D \in \TT$;
 \item otherwise, i.e. $\tip(C_M) \sqsubseteq D$ is entailed from $w$, then $p$ corresponds to the degree of belief of such inclusion of a MODIFIER in the initial knowledge base, i.e. $p \:: \ \tip(C_M) \sqsubseteq D \in \TT$.
\end{itemize}

The knowledge base obtained as the result of combining concepts $C_H$ and $C_M$ into the compound concept $C$ is called $C$-\emph{revised} knowledge base, and it is defined as follows: 
 $$\kk_C=\langle \RR, \TT \cup \{p \:: \ \tip(C) \sqsubseteq D\}, \AAA \rangle,$$  for all $D$ such that either $\tip(C_H) \sqsubseteq D$ is entailed in $w$ or $\tip(C_M) \sqsubseteq D$ is entailed in $w$, and $p$ is defined as above.
 
As an example, consider the following version of the above mentioned \emph{Pet-Fish} problem. Let KB contains the following inclusions:

\begin{quote}
    $\mathit{Fish} \sqsubseteq \mathit{LivesInWater}$ \hfill(1) \\
    $0.6 \ :: \ \tip(\mathit{Fish}) \sqsubseteq \mathit{Greyish}$ \hfill(2) \\
    $0.8 \ :: \ \tip(\mathit{Fish}) \sqsubseteq \mathit{Scaly}$ \hfill(3) \\
    $0.8 \ :: \ \tip(\mathit{Fish})  \sqsubseteq  \lnot \mathit{Affectionate}$ \hfill(4) \\
    $0.9 \ :: \ \tip(\mathit{Pet}) \sqsubseteq \lnot \mathit{LivesInWater}$ \hfill(5) \\
    $0.9 \ :: \ \tip(\mathit{Pet})  \sqsubseteq \mathit{LovedByKids}$ \hfill(6) \\ $0.9 \ :: \ \tip(\mathit{Pet})  \sqsubseteq \mathit{Affectionate}$ \hfill(7) 
\end{quote}

\noindent representing that a typical fish is greyish $(2)$, scaly $(3)$ and not affectionate $(4)$, whereas a typical pet does not live in water $(5)$, is loved by kids $(6)$ and is affectionate $(7)$. Concerning rigid properties, we have that all fishes live in water $(1)$. The logic $\pfl$ combines the concepts $\mathit{Pet}$ and $\mathit{Fish}$, by using the latter as the HEAD and the former as the MODIFIER. The prototypical Pet-Fish inherits from the prototypical fish the fact that it is scaly and not affectionate, the last one by giving preference to the HEAD since such a property conflicts with the opposite one in the modifier (a typical pet is affectionate). The scenarios in which all the three typical properties of a typical fish are inherited by the combined concept are considered as trivial and, therefore, discarded, as a consequence the property having the lowest degree ($\mathit{Greyish}$ with degree $0.6$) is not inherited. The prototypical Pet-Fish inherits from the prototypical pet only property $(6)$, since $(5)$ conflicts with the rigid property $(1)$, stating that all fishes (then, also pet fishes) live in water, whereas $(7)$ is blocked, as already mentioned, by the HEAD/MODIFIER heuristics.
Formally, the $\mathit{Pet} \sqcap \mathit{Fish}$-revised knowledge base contains, in addition to the above inclusions, the following ones:

\begin{quote}
    $0.8 \ :: \ \tip(\mathit{Pet} \sqcap \mathit{Fish}) \sqsubseteq \mathit{Scaly}$ \hfill(3') \\
    $0.8 \ :: \ \tip(\mathit{Pet} \sqcap \mathit{Fish})  \sqsubseteq  \lnot \mathit{Affectionate}$ \hfill(4') \\
    $0.9 \ :: \ \tip(\mathit{Pet} \sqcap \mathit{Fish})  \sqsubseteq \mathit{LovedByKids}$ \hfill(6') 
\end{quote}

 In \cite{lieto2018description} it has been also shown that reasoning in $\pfl$ remains in the same complexity class of standard $\alc$ Description Logics. 
 
 For the purposes of this paper, it is worth-noticing that, as mentioned, the $\pfl$ reasoning framework presented in this section has been applied, via the DEGARI system, to the generation of new compound emotions by starting from the affective ontological knowledge base named ArsEmotica. Such ontological model is described in the next section.

\section{The ArsEmotica Ontological Model enriched with the NRC Emotion Intensity Lexicon} \label{arsemotica}

The affective knowledge leveraged by the $\pfl$ logic via the DEGARI system is encoded in an ontology of emotional categories based on Plutchik’s psychological circumplex model \cite{plutchik1980general}, called ArsEmotica\footnote{The ArsEmotica ontology is available here: \url{http://130.192.212.225/fuseki/ArsEmotica-core}  and queryable via SPARQL endpoint at: \url{http://130.192.212.225/fuseki/dataset.html?tab=query&ds=/ArsEmotica-core}} and includes also concepts from the Hourglass model \cite{cambria2012hourglass}. The ontology structures emotional categories in a taxonomy, which currently includes $48$ emotional concepts. The design of the  taxonomic structure of emotional categories, of the disjunction axioms and of the object and data properties mirrors the main features of Plutchik’s circumplex model. As already mentioned, such model can be represented as a wheel of emotions (see Figure~\ref{wheel}) and encodes the following elements:

\begin{figure*}[!h]
\centering
\includegraphics[width=1\textwidth]{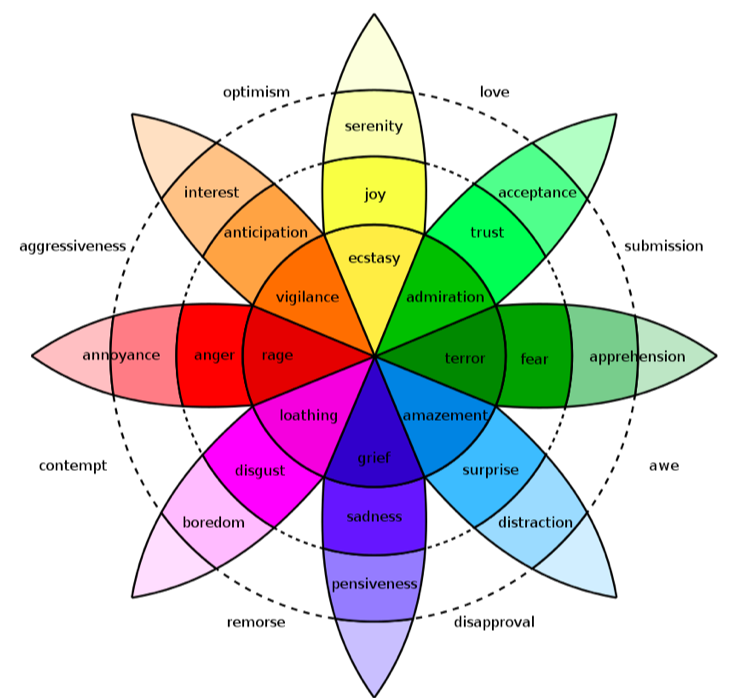}
\centering  
\caption{The Wheel of Emotion of the Plutchik Model}
\vspace{-0.2cm}
\label{wheel}
\end{figure*}

\begin{itemize}
  \item  Basic or primary emotions: {\em Joy}, {\em Trust}, {\em Fear}, {\em Surprise}, {\em Sadness}, {\em Disgust}, {\em Anger}, {\em Anticipation}; in the color wheel, this is represented by differently colored sectors.
  \item Opposites: basic emotions can be conceptualized in terms of polar opposites: {\em Joy} versus {\em Sadness}, {\em Anger} versus {\em Fear}, {\em Trust} versus {\em Disgust}, {\em Surprise} versus {\em Anticipation}.
  \item Intensity: each emotion can exist in varying degrees of intensity; in the wheel, this is represented by the vertical dimension.
  \item Similarity: emotions vary in their degree of similarity to one another; in the wheel, this is represented by the radial dimension.
  \item Complex emotions: {a complex emotion is a composition of two basic emotions; the pair of basic emotions involved in the composition is called a dyad. Looking at the Plutchik wheel, the eight emotions in the blank spaces are compositions of similar basic emotions, called primary dyads. Pairs of less similar emotions are called secondary dyads (if the radial distance between them is 2) or tertiary dyads (if the distance is 3), while opposites cannot be combined.}
\end{itemize}

We have chosen to encode the Plutchik model in the ontology for several reasons. First, it is well-grounded in psychology and general enough to guarantee a wide coverage of emotions. This is important for implementing successful strategies aimed at mapping tags to the emotional concepts of the ontology. In addition, it is easy to imagine that many different shades of emotions can be sought in the artistic domain. Literature on the psychology of art, indeed, suggests that the encoding of further complex emotions, such as {\em Pride} and {\em Shame} (belonging to the secondary and tertiary dyads, respectively, in the Plutchik model) could give further interesting results \cite{Silvia2009LookingPP}. For this reason, we considered it appropriate to broaden the classification with respect to what has been done in \cite{BertolaP16,patti2015arsemotica}, also including the compound emotions corresponding to the secondary and tertiary dyads of the Plutchik model. Second, the Plutchik wheel of emotions is perfectly compliant with the generative model underlying the $\pfl$ logic. Finally, it encodes interesting notions, e.g. emotional polar opposites, which can be exploited for finding novel, non obvious relations among artworks.

Within the ArsEmotica ontology, the class Emotion is the root for all the emotional concepts. The Emotions hierarchy includes all the $48$ emotional categories presented as distinguished labels in the model. In particular, the Emotion class has two disjoint subclasses: BasicEmotion and ComplexEmotion. Basic emotions of the Plutchik model are direct sub-classes of BasicEmotion. Each of them is specialized again into two subclasses representing the same emotion with weaker or stronger intensity (e.g. the basic emotion {\em Joy} has {\em Ecstasy} and {\em Serenity} as sub-classes). Therefore, we have $24$ emotional concepts subsumed by the BasicEmotion concept. Instead, the class CompositeEmotion has $24$ subclasses, corresponding to the primary {({\em Love, Submission, Awe, Disapproval, Remorse, Contempt, Aggressiveness e Optimism}), secondary ({\em Hope, Guilt, Curiosity, Despair, Unbelief, Envy, Cynicism e Pride}) and tertiary ({\em Anxiety, Delight, Sentimentality, Shame, Outrage, Pessimism, Morbidness, Dominance}) dyads}.
Other relations in the Plutchik model have been expressed in the ontology by means of object properties: the hasOpposite property encodes the notion of polar opposites; the hasSibling property encodes the notion of similarity and the isComposedOf property encodes the notion of composition of basic emotions. Moreover, a data type property hasScore was introduced to link each emotion with an intensity value i mapped into the above mentioned Hourglass model.  

The devised model allows attributing complex emotions from basic ones by exploiting simple SWRL rules (i.e. if-then clauses) allowing to infer, via the isComposedOf property connecting Basic and Composite Emotions,  the fact that if an agent feels two emotions (suppose for a given item), and if these emotions jointly constitute a Composite Emotion, then the latter emotion will be automatically assigned to the agent in order to better describe his/her aesthetic experience.

Due to the need of modeling the links between words in a language and the emotions they refer to, the ArsEmotica Ontology is also integrated with the ontology framework LExicon Model for ONtologies (LEMON) \cite{mccrae2011linking}. In particular, such integration allows   differentiating explicitly between the language level (lexicon-based) and the conceptual one in representing the emotional concepts \cite{patti2015arsemotica}. Within this enriched framework, it is possible to associate a plethora of emotional words, with the encoding of language information, to the corresponding emotional concepts. In this work, we have used the ArsEmotica model of emotional concepts with the NRC Emotion Intensity Lexicon mentioned above \cite{mohammad2017word}. Such lexicon provides a list of English words, each with real-values representing intensity scores for the eight basic emotions of Plutchik's theory. The lexicon includes close to $10,000$ words including terms already known to be associated with emotions as well as terms that co-occur in Twitter posts that convey emotions. The intensity scores were obtained via crowdsourcing, using best-worst scaling annotation scheme. For our purposes, we considered the most frequent terms available in such lexicon (and associated to the basic emotions of the Plutchik wheel) as typical features of such emotions. In this way, once the prototypes of the basic emotional concepts were formed, the $\pfl$ reasoning framework was used to generate the compound emotions.

\section{DEGARI: Generating Novel Emotions from ArsEmotica}\label{degari1}
In this section we describe DEGARI, the system exploiting the logic $\pfl$ on the ArsEmotica knowledge base in order to generate and suggest novel emotion related contents and tested on the  RaiPlay catalog \footnote{\url{https://www.raiplay.it}}, as well as on two artwork datasets: WikiArt Emotions\footnote{\url{http://saifmohammad.com/WebPages/wikiartemotions.html}} and ArsMeteo \cite{acotto2009arsmeteo,BertolaP16}. DEGARI is  implemented in Python and it makes use of the library \texttt{owlready2}\footnote{\url{https://pythonhosted.org/Owlready2/}} for relying on the  services of efficient DL reasoners (like HermiT).

 DEGARI's prototypes generation proceeds in two steps: in the first one, it  builds a prototypical description of basic emotions in the language of the logic $\pfl$, in order to describe their typical properties; as a second step, it exploits the above described reasoning mechanism of such a Description Logic in order to combine the prototypical descriptions of pairs of basic emotions, generating the prototypical description of compound emotions.
 As mentioned above, the obtained ontology is then tested by re-classifying the items belonging to RaiPlay, Wiki Art and ArsMeteo, keeping the generated compound emotions into account: this allows us to describe a novel and completely explainable recommending system, which is able to suggest items belonging also to compound emotions.

Concerning the first step, DEGARI builds a knowledge base in the logic $\pfl$ characterized by typicality inclusions of the form
$$p \ :: \ \tip(\mathit{BasicEmotion}) \sqsubseteq \mathit{Property}$$ where $\mathit{BasicEmotion}$ is one of the eight basic emotions of the  Plutchik model: Joy, Trust, Fear, Surprise, Sadness, Disgust, Anger, and Anticipation. Typical properties are selected from the list of words characterizing each basic emotion in the NRC Emotion Intensity Lexicon where, as already mentioned, the probability $p$ represents the intensity score for the emotion. In detail, for each basic emotion, we consider the six properties/words having the highest scores.

 
     As an example, consider the basic emotion $\mathit{Joy}$. The words having the highest scores are happiness ($0.98$), bliss ($0.97$), to celebrate ($0.97$), jubilant ($0.97$), ecstatic ($0.95$), and euphoria ($0.94$). Therefore, the knowledge base generated by DEGARI will contain, among others, the following inclusions:
    \begin{quote}
    $\mathit{Joy} \sqsubseteq \lnot \mathit{Holocaust}$ \\
        $0.98 \ :: \ \tip(\mathit{Joy}) \sqsubseteq \mathit{Happiness}$ \\
        $0.97 \ :: \ \tip(\mathit{Joy}) \sqsubseteq \mathit{Bliss}$ \\
        $0.97 \ :: \ \tip(\mathit{Joy}) \sqsubseteq \mathit{Celebrating}$ \\
        $0.97 \ :: \ \tip(\mathit{Joy}) \sqsubseteq \mathit{Jubilant}$
\\
        $0.95 \ :: \ \tip(\mathit{Joy}) \sqsubseteq \mathit{Ecstatic}$
    \\
        $0.94 \ :: \ \tip(\mathit{Joy}) \sqsubseteq \mathit{Elation}$
         \end{quote}

     DEGARI then computes novel compound emotions by combining existing ones (by using the same logical procedure of the \emph{pet-fish} problem). As an example, let us consider the combination of the above basic emotion $\mathit{Joy}$ with $\mathit{Fear}$, whose prototypical description is as follows:

     \begin{quote}
         $ 0.96 \ :: \ \tip(\mathit{Fear}) \sqsubseteq \mathit{Kill}$ \\
         $ 0.95 \ :: \ \tip(\mathit{Fear}) \sqsubseteq \mathit{Annihilate}$ \\
         $ 0.95 \ :: \ \tip(\mathit{Fear}) \sqsubseteq \mathit{Terror}$ \\
         $ 0.98 \ :: \ \tip(\mathit{Fear}) \sqsubseteq \mathit{Torture}$ \\
         $ 0.97 \ :: \ \tip(\mathit{Fear}) \sqsubseteq \mathit{Terrorist}$ \\
         $ 0.97 \ :: \ \tip(\mathit{Fear}) \sqsubseteq \mathit{Horrific}$
     \end{quote}

     In order to obtain a description of the compound emotion $\mathit{Guilt}$ as the result of the combination of the two basic emotions ($\mathit{Joy} \sqcap \mathit{Fear}$) in the logic $\pfl$, DEGARI combines the two basic emotions by implementing a variant of CoCoS \cite{cilc2018}, a Python implementation of reasoning services for the logic $\pfl$ in order to exploit efficient DLs reasoners for checking both the consistency of each generated scenario and the existence of conflicts among properties, following the line of the system DENOTER \cite{ecainostro}. More in detail, DEGARI considers both the available choices for the HEAD and the MODIFIER, and it allows restricting its concern to a given and fixed number of inherited properties. The combined emotion $\mathit{Guilt}$ has the following $\pfl$ description (concept $\mathit{Joy} \sqcap \mathit{Fear}$):
 
  \begin{quote}
         $ 0.98 \ :: \ \tip(\mathit{Joy} \sqcap \mathit{Fear}) \sqsubseteq \mathit{Happiness}$ \\
         $ 0.97 \ :: \ \tip(\mathit{Joy} \sqcap \mathit{Fear}) \sqsubseteq \mathit{Celebrating}$ \\
         $ 0.97 \ :: \ \tip(\mathit{Joy} \sqcap \mathit{Fear}) \sqsubseteq \mathit{Bliss}$ \\
         $ 0.98 \ :: \ \tip(\mathit{Joy} \sqcap \mathit{Fear}) \sqsubseteq \mathit{Torture}$ \\
         $ 0.97 \ :: \ \tip(\mathit{Joy} \sqcap \mathit{Fear}) \sqsubseteq \mathit{Terrorist}$ \\
         $ 0.97 \ :: \ \tip(\mathit{Joy} \sqcap \mathit{Fear}) \sqsubseteq \mathit{Horrific}$
    \end{quote}

     Obviously, rigid properties of basic emotions (if any) are inherited by the compound emotion (in the example, $\mathit{Joy} \sqcap \mathit{Fear} \sqsubseteq \lnot \mathit{Holocaust}$), and this retain the system from considering any inconsistent typical properties even if they have the highest probability.

It is worth noticing that the properties of the derived emotion are still expressed in the language of the logic $\pfl$, therefore the combined emotion, $\mathit{Guilt}$  in the example, can be further combined with another emotion, in order to iterate the procedure.

As mentioned above, the DEGARI component devoted to compute the above described concept combination of prototypical descriptions relies on CoCoS \cite{cilc2018}, an implementation of a reasoning machinery for the logic $\pfl$, generating scenarios and choosing the selected one(s) as described in Section \ref{logic}. CoCoS is implemented in Python and exploits the translation of an $\alctr$ knowledge base into standard $\alc$ introduced in \cite{AIJ2014} and adopted by the system RAT-OWL \cite{ratowl}. CoCoS makes use of the above mentioned library \texttt{owlready2}, which allows relying  on the  services of efficient DL reasoners, e.g. the HermiT reasoner. 

 CoCoS is embedded in DEGARI and allows one: i) to include the logical descriptions of the concepts to be combined; ii) to select which among the concepts has to be intended as HEAD and as MODIFIER(s); iii) to choose how many typical properties one wants to inherit in the scenarios that will be selected by $\pfl$. In addition to presenting the selected scenario with typical properties of the combined concept, CoCoS also allows the users to select alternative scenarios, ranging from more trivial to more surprising ones.

\section{Reclassification of Emotion-related Content based on DEGARI}  \label{degari2} 

By starting from the generated prototypes of the compound emotions in ArsEmotica, DEGARI is also able to perform an emotion-oriented reclassification of the items of the considered datasets.  

In particular, DEGARI employs two different strategies to extract metadata from the items to reclassify. 
In a first case (e.g., for the ArsMeteo and WikiArt datasets) the metadata are either stored in the provided resource (e.g., in WikiArt) or are the result of a social tagging activity based on the artistic community. In the second case (e.g., in the case of the RaiPlay dataset) the metadata associated to every and each item (title, name of the program/episode, description of the program/series, description of the episode) are extracted from a crawler. Such metadata are then used to generate the typical description of the items via the computation of the most frequent terms retrieved in their textual description (the assumption is that the most frequently used terms to describe an item are also the ones that are more typically associated to them). The frequencies are computed as the proportion of each property with respect to the set of all properties characterizing the item, in order to compare them with the properties of the derived emotion. If the item
     contains all the rigid properties and at least the $30\%$ of the typical properties of the compound emotion under consideration, then the item is classified as belonging to it. 
     Last, DEGARI suggests the set of classified contents, in a descending order of compatibility, where the rank of compatibility of a single item with respect to an emotion is intuitively obtained as the sum of the frequencies of ``compatible'' concepts, i.e. concepts belonging to both the item and the prototypical description of the genre. 
    Formally:
    
    \begin{definition}
        Given an item $m$, let $\mathit{DerivedEmotion}$ be a compound emotion generated from the ArsEmotica mode as defined in Section \ref{degari1} and let $\mathcal{S}_m$ be the set of words occurring in $m$. Given a knowledge base KB of compound emotions built by DEGARI, we say that $m$ is compatible with $\mathit{DerivedEmotion}$ if the following conditions hold:
        \begin{enumerate}
            \item $m$ contains all rigid properties of $\mathit{DerivedEmotion}$, i.e. $\{ C \mid \mathit{DerivedEmotion} \sqsubseteq C \in \ \mbox{KB} \} \subseteq \mathcal{S}_m$
            \item $m$ contains at least the $30\%$ of typical properties of $\mathit{DerivedEmotion}$ , i.e. $$\frac{\mid \mathcal{S}_m \cap \mathcal{S}_{\mathit{DerivedEmotion}} \mid}{\mid \mathcal{S}_{\mathit{DerivedEmotion}} \mid} \geq 0.3,$$ where $\mathcal{S}_{\mathit{DerivedEmotion}}$ is the set of typical properties of $\mathit{DerivedEmotion}$.
        \end{enumerate}
    \end{definition}

 \noindent    As another example, consider the  derived emotion $\mathit{Joy} \ \sqcap \ \mathit{Surprise}$, which in the Plutchik wheel corresponds to the combined emotion ``delight". The knowledge base in the logic $\pfl$ describing such a compound emotion is  as follows:
     
    \begin{quote}
         $ 0.98 \ :: \ \tip(\mathit{Joy} \sqcap \mathit{Surprise}) \sqsubseteq \mathit{Happiness}$ \\
         $ 0.97 \ :: \ \tip(\mathit{Joy} \sqcap \mathit{Surprise}) \sqsubseteq \mathit{Bliss}$ \\
         $ 0.97 \ :: \ \tip(\mathit{Joy} \sqcap \mathit{Surprise}) \sqsubseteq \mathit{Celebrating}$ \\
         $ 0.97 \ :: \ \tip(\mathit{Joy} \sqcap \mathit{Surprise}) \sqsubseteq \mathit{Jubilant}$\\ 
         $ 0.95 \ :: \ \tip(\mathit{Joy} \sqcap \mathit{Surprise}) \sqsubseteq \mathit{Estatic}$ \\
         $\mathit{Joy} \sqcap \mathit{Surprise} \sqsubseteq \lnot \mathit{Holocaust}$ \\
                  $\mathit{Joy} \sqcap \mathit{Surprise} \sqsubseteq \lnot \mathit{Anticipation}$
         
    \end{quote}

\begin{figure*}[!h]
\centering
\includegraphics[angle=270,origin=c,
width=1\textwidth]{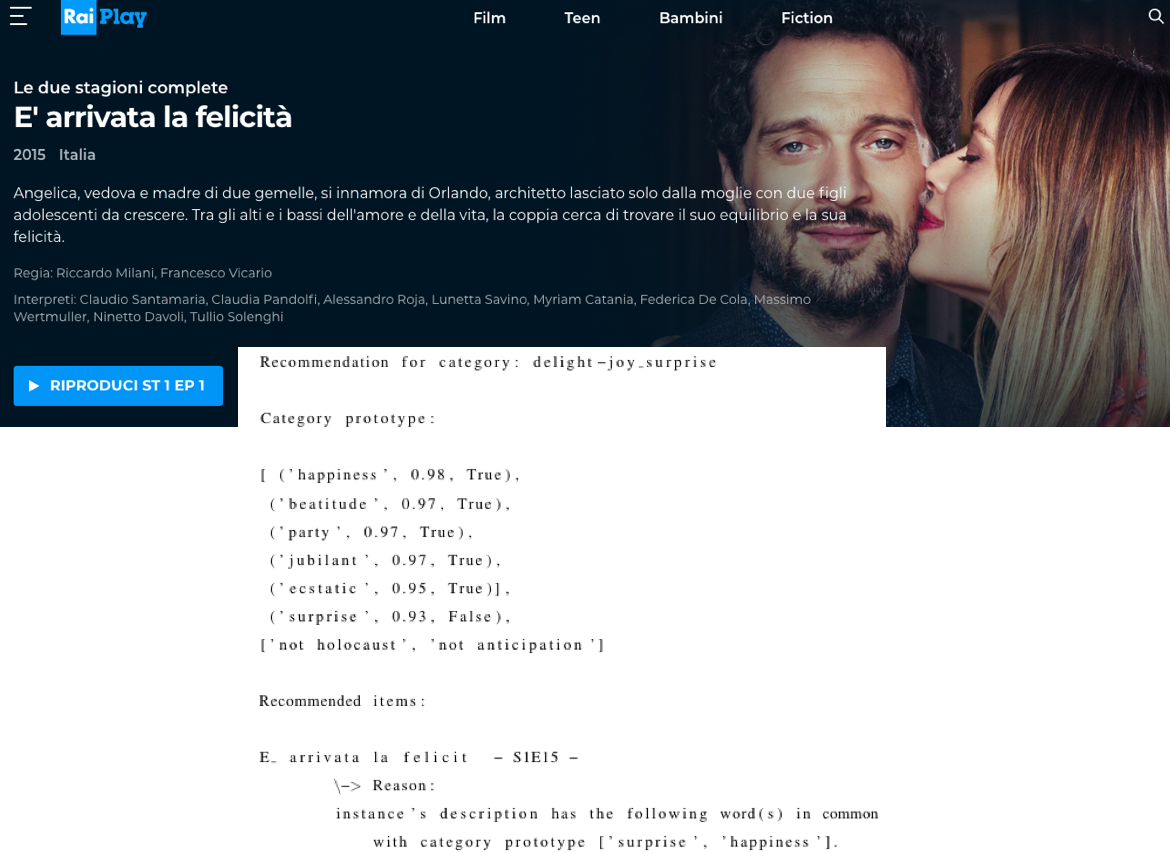}
\centering  
\caption{An Example of the Laconic Explanation provided by DEGARI about the reasons determining the reclassification of an item in the RaiPlay dataset.}
\label{explanation}
\end{figure*}

\noindent  For instance, the multimedia item ``È arrivata la felicità"  (``Happiness has arrived")  (\url{https://www.raiplay.it/programmi/earrivatalafelicita})  from the RaiPlay dataset is reclassified in the novel, generated emotion $\mathit{Joy} \ \sqcap \ \mathit{Surprise}$, since:

     \begin{itemize}
         \item all rigid properties of both basic emotions are satisfied, that is to say neither $\mathit{Holocaust}$ nor $\mathit{Anticipation}$ belong to the properties extracted for the item;
         \item more than the $30\%$ of the typical properties\footnote{The $30\%$ threshold was empirically determined: i.e., it is the percentage that provides the better trade-off between overcategorization and missed categorizations.} of the compound emotion are satisfied; in particular, ``È arrivata la felicità'' has $\mathit{Happiness}$ ($0.98$) and $\mathit{Surprise}$ ($0.93$). 
     \end{itemize}
     This item will be then recommended by DEGARI as shown in Figure \ref{explanation}.


\section{Explanation}\label{explsection}

Figure \ref{explanation} also shows how DEGARI can be considered as an explainable AI system: indeed, an explanation of the reasons why the multimedia item ``È arrivata la felicità"  (``Happiness has arrived") has been reclassified in the compound concept is provided, in order to let the user be aware of the procedure of  the system. As a difference with ``black box'' approaches, DEGARI explicitly reports that the ``instance's description has the following word(s) in common with category prototype'', followed by the two matching properties {\em Surprise} and {\em Happiness}. The list of matching properties provided by DEGARI as a textual output is generated during the reclassification process. For each considered artwork, the properties of the artwork's prototype's are compared to the ones of the emotion's prototype. The reclassification is based only on the matching properties (if present), therefore providing an explanation of the  reclassification is as simple as storing the matches in an array and printing it as an output. 
Moreover, the whole procedure is completely transparent and could be used to further expand the feedback provided by system: from the axiom system (and the corresponding semantics of the Description Logic with typicality) to the DISPONTE semantics adopted by the logic $\pfl$ in order to compute the prototypical description of the compound emotion.

\section{Evaluation}\label{evaluation}

DEGARI evaluation has been carried out on three different datasets and evaluated in a fourfold way. 
The datasets considered (described in detail below) are: ArsMeteo, the RaiPlay catalog, and Wiki Art Emotion. The evaluation has concerned a first, completely automatic, test consisting in calculating the percentage of the reclassified items within the novel hybrid emotion classes generated by the system via $\pfl$. 
In this case, the spread of the reclassified items along most of the concepts of the wheel of emotion has been considered as a positive indicator. This aspect, indeed, shows how the created prototypes of the compound emotions are mostly meaningful and able to reclassify the artistic content available in the three original datasets.
 A second evaluation has concerned the use of an ablation experiment testing the components of the $\pfl$ logic used by DEGARI in order to determine, if any, a difference in terms of the obtained output. A third evaluation has consisted in testing, within the pipeline provided by DEGARI, two different emotional lexicons: the above mentioned NRC lexicon and the lexicon of Shaver's model of emotions \cite{shaver1987emotion}. The rationale of this experiment was to assess to what extent the combinatorial mechanisms used by DEGARI were dependent from the particular lexicons used. 
A fourth evaluation, finally, aimed at measuring the satisfaction of the potential users of the system when exposed to the contents of the novel categories suggested by DEGARI, consisted in a user study\footnote{This is one of the most commonly used methodology for the evaluation of recommender systems based on controlled small groups analysis, see \cite{shani2011evaluating}.} involving $44$ subjects, who  
evaluated a total of $30$ recommendations generated by the system.  All the participants were recruited online using an availability sampling strategy. Participants were all naive to the experimental procedure and to the aims of the study.

In the following, we briefly describe the adopted datasets: two of them are art-related ones (ArsMeteo and WikiArt Emotion), while the RaiPlay dataset contains all the multimedia items (e.g. movies, tv series, tv shows, documentaries etc.) available on the online multimedia platform or RAI, Radiotelevisione Italiana.

\subsection{ArsMeteo Dataset}

ArsMeteo \cite{acotto2009arsmeteo} is an art portal for sharing artworks and their emerging, connective meanings. Its development is led by a non-profit cultural organization called Associazione Culturale ArsMeteo (AMA), based in Turin, Italy. It enables the collection of digital (or digitalized) artworks and performances, belonging to a variety of artistic forms including poems, videos, pictures and music. Meanings are given by the tagging activity of the community. All contents are accessible as ``digital commons". Currently, the portal has collected over 350,000 visits and gathered a collection of over $9,000$ artifacts produced by $307$ artists; it has collected almost $38,000$ tags.

\subsection{RAIPLAY Dataset}

The RaiPlay dataset is composed of $4,612$ multimedia items extracted from RaiPlay \url{https://www.raiplay.it/}, the online platform of RAI, the national Italian broadcaster. Such dataset contains different types of multimedia content grouped in six main narrative categories: Movies, Fiction, Kids, TV Series, Drama, Comedy. 
As mentioned, each multimedia item/episode is equipped with metadata, including: title, name of the program/episode, description of the program/series, description of the episode. Such descriptions are used by DEGARI to extract the relevant information to associate to every item and to decide whether, given the extracted information, the item should be reclassified in one of the previously generated compound emotions.

\subsection{WikiArt Emotions}

WikiArt Emotions is a dataset of 4,105 artworks with annotations for the emotions evoked in the observer \cite{mohammad-kiritchenko-2018-wikiart}. The artworks were selected from the online visual art encyclopedia WikiArt.org. Each piece of art is annotated for one or more of $20$ emotion categories (including neutral).
Annotations were obtained via crowdsourcing, asking annotators to include all the emotions evoked by the title of the artwork, the image of the artwork or the artwork as a whole. The annotators were also asked to specify if the artwork depicted a face or a human body (but not a face): this additional information is included in the dataset (if an artwork didn't depict a face nor a body, it was marked as ``none").
In order to decide if an emotion applies or not to an artwork, the authors specified an aggregation threshold of $40\%$: if at least $40\%$ of the responses indicated that a certain emotion applied, then the label was chosen. Other distributions on the dataset with different aggregation thresholds ($30\%$ and $50\%$) are available, but we chose to use the $40\%$ threshold version, as recommended by the authors of the dataset \cite{mohammad-kiritchenko-2018-wikiart}.

\vspace{8pt}
\begin{footnotesize}
\begin{longtable}{l|r r|r r|r r}
\toprule
\rule[-1ex]{0pt}{2.5ex}  & \multicolumn{2}{p{2.1cm}|}{\textbf{ArsMeteo\newline(9171 artworks)}}  & \multicolumn{2}{p{2.1cm}|}{\textbf{RaiPlay\newline(4612 media items)}} & \multicolumn{2}{p{2.1cm}}{\textbf{WikiArt \newline Emotions\newline(4105 artworks)}}  \\ 
\midrule
\rule[-1ex]{0pt}{2.5ex} {\textbf{Emotion}} & \multicolumn{1}{b{0.8cm}}{\textbf{\scriptsize  Reclassified\newline Items}} & \multicolumn{1}{b{0.9cm}|}{\textbf{\scriptsize Percentage}} & \multicolumn{1}{b{0.8cm}}{\textbf{\scriptsize  Reclassified\newline Items}} & \multicolumn{1}{b{0.9cm}|}{\textbf{\scriptsize Percentage}} & \multicolumn{1}{b{0.8cm}}{\textbf{\scriptsize Reclassified\newline Items}} & \multicolumn{1}{b{0.9cm}}{\textbf{\scriptsize Percentage}} \\ 
\midrule
\rule[-1ex]{0pt}{2.5ex} {\textbf{aggressiveness (anger\_anticipation)}} & $12$ & $0.13\%$ & $7$ & $0.15\%$ & $0$ & $0.00\%$ \\ 
\midrule
\rule[-1ex]{0pt}{2.5ex} {\textbf{aggressiveness (anticipation\_anger)}} & $11$ & $0.12\%$ & $2$ & $0.04\%$ & $0$ & $0.00\%$  \\ 
\midrule
\rule[-1ex]{0pt}{2.5ex} {\textbf{anxiety (anticipation\_fear)}} & $33$ & $0.36\%$ & $88$ & $1.91\%$ & $0$ & $0.00\%$  \\ 
\midrule
\rule[-1ex]{0pt}{2.5ex} {\textbf{awe (fear\_surprise)}} & $63$ & $0.69\%$ & $132$ & $2.86\%$ & $508$ & $12.38\%$  \\ 
\midrule
\rule[-1ex]{0pt}{2.5ex} {\textbf{contempt (anger\_disgust)}} & $13$ & $0.14\%$ & $2$ & $0.04\%$ & $0$ & $0.00\%$  \\
\midrule
\rule[-1ex]{0pt}{2.5ex} {\textbf{curiosity (surprise\_trust)}} & $47$ & $0.51\%$ & $49$ & $1.06\%$ & $508$ & $12.38\%$  \\
\midrule
\rule[-1ex]{0pt}{2.5ex} {\textbf{cynicism (anticipation\_disgust)}} & $5$ & $0.05\%$ & $0$ & $0.00\%$ & $0$ & $0.00\%$  \\ 
\midrule
\rule[-1ex]{0pt}{2.5ex} {\textbf{delight (joy\_surprise)}} & $52$ & $0.57\%$ & $117$ & $2.54\%$ & $1672$ & $40.73\%$  \\
\midrule
\rule[-1ex]{0pt}{2.5ex} {\textbf{despair (fear\_sadness)}} & $19$ & $0.21\%$ & $20$ & $0.43\%$ & $0$ & $0.00\%$  \\
\midrule
\rule[-1ex]{0pt}{2.5ex} {\textbf{disapproval (sadness\_surprise)}} & $58$ & $0.63\%$ & $64$ & $1.39\%$ & $404$ & $9.84\%$  \\ 
\midrule
\rule[-1ex]{0pt}{2.5ex} {\textbf{dominance (anger\_trust)}} & $11$ & $0.12\%$ & $2$ & $0.04\%$ & $0$ & $0.00\%$  \\ 
\midrule
\rule[-1ex]{0pt}{2.5ex} {\textbf{envy (anger\_sadness)}} & $25$ & $0.27\%$ & $21$ & $0.46\%$ & $0$ & $0.00\%$  \\
\midrule
\rule[-1ex]{0pt}{2.5ex} {\textbf{envy (sadness\_anger)}} & $17$ & $0.19\%$ & $20$ & $0.43\%$ & $0$ & $0.00\%$  \\ 
\midrule
\rule[-1ex]{0pt}{2.5ex} {\textbf{guilt (fear\_joy)}} & $22$ & $0.24\%$ & $73$ & $1.58\%$ & $1618$ & $39.42\%$  \\
\midrule
\rule[-1ex]{0pt}{2.5ex} {\textbf{guilt (joy\_fear)}} & $24$ & $0.26\%$ & $73$ & $1.58\%$ & $1618$ & $39.42\%$  \\
\midrule
\rule[-1ex]{0pt}{2.5ex} {\textbf{hope (anticipation\_trust)}} & $50$ & $0.55\%$ & $51$ & $1.11\%$ & $661$ & $16.10\%$  \\
\midrule
\rule[-1ex]{0pt}{2.5ex} {\textbf{hope (trust\_anticipation)}} & $44$ & $0.48\%$ & $52$ & $1.13\%$ & $661$ & $16.10\%$  \\
\midrule
\rule[-1ex]{0pt}{2.5ex} {\textbf{love (joy\_trust)}} & $24$ & $0.26\%$ & $72$ & $1.56\%$ & $1618$ & $39.42\%$  \\
\midrule
\rule[-1ex]{0pt}{2.5ex} {\textbf{morbidness (disgust\_joy)}} & $25$ & $0.27\%$ & $70$ & $1.52\%$ & $1618$ & $39.42\%$  \\
\midrule
\rule[-1ex]{0pt}{2.5ex} {\textbf{optimism (anticipation\_joy)}} & $23$ & $0.25\%$ & $24$ & $0.52\%$ & $1413$ & $34.42\%$  \\
\midrule
\rule[-1ex]{0pt}{2.5ex} {\textbf{outrage (anger\_surprise)}} & $41$ & $0.45\%$ & $46$ & $1.00\%$ & $508$ & $12.38\%$  \\
\midrule
\rule[-1ex]{0pt}{2.5ex} {\textbf{pessimism (anticipation\_sadness)}} & $37$ & $0.40\%$ & $20$ & $0.43\%$ & $412$ & $10.04\%$  \\
\midrule
\rule[-1ex]{0pt}{2.5ex} {\textbf{pessimism (sadness\_anticipation)}} & $29$ & $0.32\%$ & $20$ & $0.43\%$ & $0$ & $0.00\%$  \\
\midrule
\rule[-1ex]{0pt}{2.5ex} {\textbf{pride (anger\_joy)}} & $20$ & $0.22\%$ & $72$ & $1.56\%$ & $1618$ & $39.42\%$  \\
\midrule
\rule[-1ex]{0pt}{2.5ex} {\textbf{remorse (disgust\_sadness)}} & $31$ & $0.34\%$ & $20$ & $0.43\%$ & $0$ & $0.00\%$  \\
\midrule
\rule[-1ex]{0pt}{2.5ex} {\textbf{sentimentality (sadness\_trust)}} & $18$ & $0.20\%$ & $20$ & $0.43\%$ & $0$ & $0.00\%$  \\
\midrule
\rule[-1ex]{0pt}{2.5ex} {\textbf{sentimentality (trust\_sadness)}} & $28$ & $0.31\%$ & $20$ & $0.43\%$ & $0$ & $0.00\%$  \\
\midrule
\rule[-1ex]{0pt}{2.5ex} {\textbf{shame (disgust\_fear)}} & $35$ & $0.38\%$ & $88$ & $1.91\%$ & $0$ & $0.00\%$  \\
\midrule
\rule[-1ex]{0pt}{2.5ex} {\textbf{submission (fear\_trust)}} & $33$ & $0.36\%$ & $88$ & $1.91\%$ & $0$ & $0.00\%$  \\
\midrule
\rule[-1ex]{0pt}{2.5ex} {\textbf{unbelief (disgust\_surprise)}} & $35$ & $0.38\%$ & $44$ & $0.95\%$ & $508$ & $12.38\%$  \\
\midrule
\rule[-1ex]{0pt}{2.5ex} {\textbf{unbelief (surprise\_disgust)}} & $47$ & $0.51\%$ & $47$ & $1.02\%$ & $508$ & $12.38\%$  \\
\midrule
\rule[-1ex]{0pt}{2.5ex} {\textbf{OVERALL${}^\star$}} & $166$ & $1.81\%$ & $235$ & $5.49\%$ & $2434$ & $59.29\%$  \\
\bottomrule
\caption{Automatic evaluation overall data. ${}^\star$ \emph{This is NOT a sum: the overall count represents the total number of artworks classified for at least one emotion.}} 

\label{tabellaoverall}
\end{longtable}
\end{footnotesize}

\subsection{Automatic Reclassification}

The obtained results for what concerns the automatic evaluation are presented in Table \ref{tabellaoverall}. 
Overall, the figure shows that for two of the three datasets (ArsMeteo and RaiPlay) DEGARI is able to reclassify and spread the original items along the entire wheel of emotions assumed by the Plutchik model, thus allowing a more fine grained characterization.

In these two cases, the percentage of the reclassified items is of $5.49\%$  and $1.81\%$, respectively. On the other hand, the WikiArt Emotions dataset contains orthogonal results since, in this case, $16$ out of the $31$ generated compound emotions are filled with reclassified items. In this case, however, a large part of the dataset  ($59.29\%$) is involved in such a reclassification.

\begin{figure*}[!h]
\centering
\includegraphics[width=1.1\textwidth]{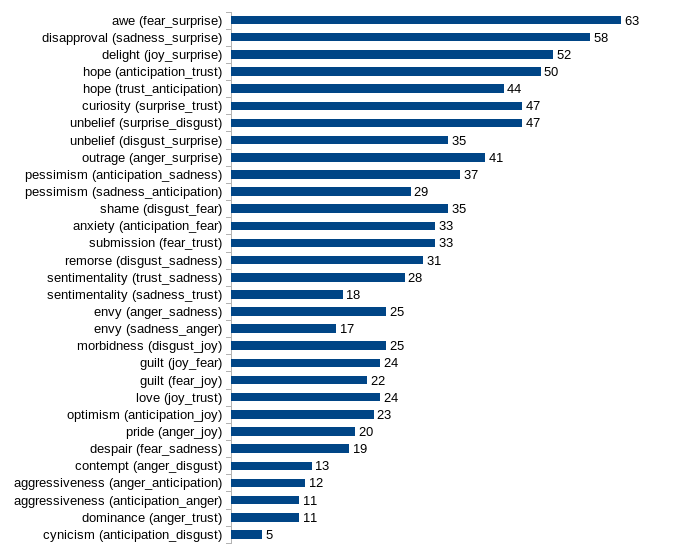}
\centering  
\caption{Top Reclassified Content of ArsMeteo in the Compound Emotional Classes generated by DEGARI}
\vspace{-0.2cm}
\label{arsmeteo}
\end{figure*}

The main reason for these orthogonal results is in the kind of input considered by DEGARI. Indeed, while for ArsMeteo and RaiPlay the metadata associated to the items are either the result of a social tagging activity by a community of artists (like in ArsMeteo\footnote{The ArsMeteo dataset has the additional difficulty of being a heterogeneous dataset, touching different artistic genres (from poetry to literature, to paintings).}) or the result of an information extraction pipeline (in RaiPlay); in WikiArt Emotions, all the metadata associated to the items are the result of a controlled crowdsourcing activity based on predefined emotion tags. 
\begin{figure*}[!h]
\centering
\includegraphics[width=1.1\textwidth]{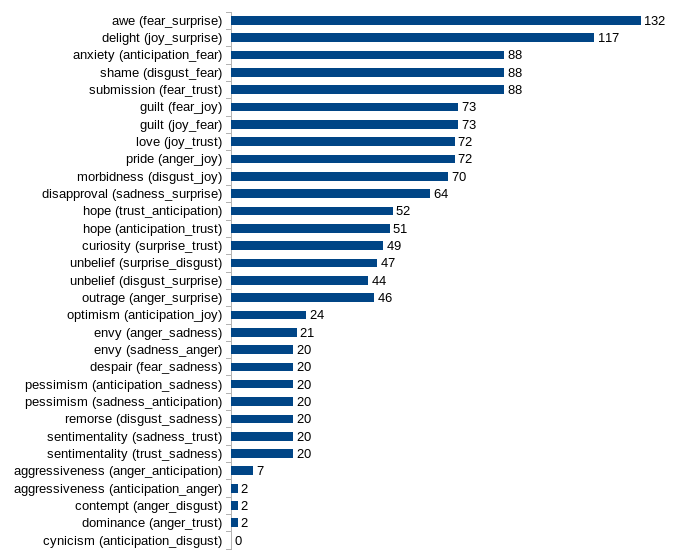}
\centering  
\caption{Top Reclassified Content of RaiPlay in the Compound Emotional Classes generated by DEGARI}
\vspace{-0.2cm}
\label{raiplay}
\end{figure*}

\begin{figure*}[!h]
\centering
\includegraphics[width=1.1\textwidth]{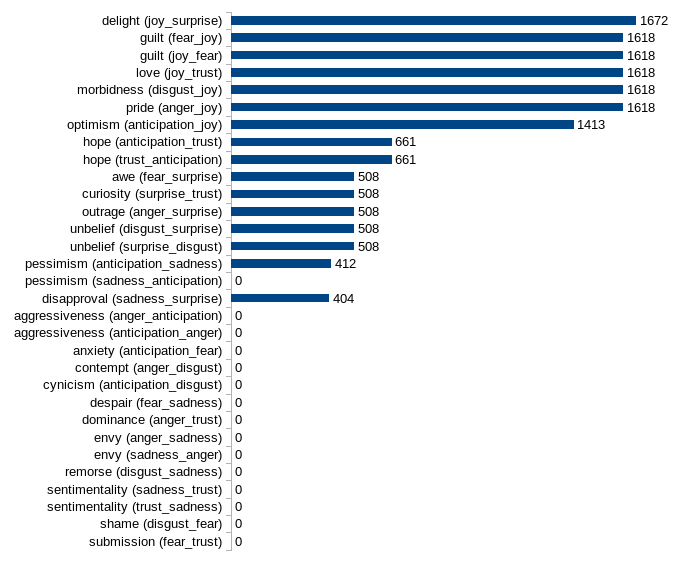}
\centering  
\caption{Top Reclassified Content of WikiArt Emotions in the Compound Emotional Classes generated by DEGARI}
\vspace{-0.2cm}
\label{wikiart}
\end{figure*}

While this fact, on one hand, creates - for the WikiArt Emotions dataset - cleaner metadata and allows the reclassifications of most of the items available in the dataset, on the other hand, it forces the user to use a predefined vocabulary for annotation that, as such, inhibits more free associations that could have led to a wider reclassification and redistribution of the items along the entire Plutchik wheel. 
In all the $3$ cases, however, most of the compound emotions generated by DEGARI are filled with new items.

 In order to test the efficacy of the method employed by the DEGARI system (relying, as mentioned, on the $\pfl$ logic) we conducted an ablation experiment in order to determine the components of  $\pfl$ shaping the above presented output.  Although the adoption of such an experimental setting is not usual in symbolic approaches  (since the causal connection between the output and the composing elements of a formalism or of an algorithms is - in such approaches - directly made explicit)\footnote{This aspect represents an important difference with the currently \emph{in auge} deep learning models that, on the other hand, suffer from the well known opacity issues and, as a consequence, require by default the adoption of such experimental technique in order to investigate which elements of such models are the relevant ones for producing and justifying a particular output.}, the dissection imposed by such a setting allowed, nonetheless, a better understanding of what are the mandatory and the corollary elements of the  formalism upon which our system is based on. In particular, as described in section \ref{logic}, $\pfl$ relies on three  ingredients: a Description Logic of Tipicality $\alct$, the DISPONTE semantics, and the Head/Modifier heuristics. The  first two ingredients (i.e. the $\alct$ and the  DISPONTE logics) are mandatory elements in order to obtain the commonsense conceptual combination proposed in $\pfl$ and therefore cannot be ``ablated" (since no output would be provided in case of deleting either the typical component of the logic or the probabilistic one). On the other hand, the head/modifier heuristics is an additional element built on the top of the probabilistic commonsense framework and, as such, can be an element subject to an ablation study. 
We performed such a study and the results are available below. The Table \ref{t:tags3} shows the evident advantage (in terms of reclassified items) of using this heuristics compared to the cases in which it is not used.

In the figures \ref{arsmeteo}, \ref{raiplay}, \ref{wikiart}, finally, are also reported, for each dataset, the generated compound emotions that have received more reclassifications (the horizontal histograms indicate the number of the reclassified items for each compound emotion).


\begin{table*}
\label{ablation}
\caption{Ablation Experiment of DEGARI showing the difference of considering (or not) the Head/Modifier (H/M) Cognitive Heuristics and its crucial effect on emotion reclassifications}
\label{t:tags3}
\centering
\begin{tabular}{c|c|c|c }
\toprule
\rule[-1ex]{0pt}{2.5ex}  & {\textbf{ArsMeteo}}  & {\textbf{WikiArt}} & {\textbf{RaiPlay}}  \\ 
\midrule
\rule[-1ex]{0pt}{2.5ex} {\textbf{Total (with H/M) }} & $607$ & $1428$ & $15853$  \\ 
\midrule
\rule[-1ex]{0pt}{2.5ex} {\textbf{ Total (without H/M)}} & $212$ & $570$ & $6531$  \\ 
\midrule
%
\rule[-1ex]{0pt}{2.5ex} {\textbf{ Delta (n. of reclassifications)}} & $-395$ & $-854$ & $-9322$  \\ 
\midrule
\rule[-1ex]{0pt}{2.5ex} {\textbf{Delta (percentage of reclassifications)}} & $-65.07$\% & $-59.97$\% & $-58.80$\%  \\ 
\bottomrule
\end{tabular} 
\vspace{7pt}
\end{table*}

\subsubsection{Extended Experiment with the Shaver Lexicon}


In order to extend our experimental evaluation, we decided to test the well known lexicon provided by Shaver's emotion model \cite{shaver1987emotion} (along with the Plutchik model and with the commonsense compositional mechanisms adopted by DEGARI over it) and compare the reclassifications/recommendations provided by our system with the first evaluation (using the NRC lexicon described above). Created with the goal of investigating the intuitions behind people's conceptualization of emotions -- as a way to reveal their nature --, Shaver's model has been created through a hierarchical clustering analysis of lexical data, having the Rosch's theory of prototypes as reference \cite{rosch75cognitive}. In other words: the work by Shaver and colleagues investigates the role of prototypes in representing emotions in human conceptualization. It uses, as typical features of each emotional concept, the most common words used to described them \footnote{We remind here that, according to Rosch's theory \cite{rosch75cognitive}, the prototypes are commonsense representations of a given concept. Here concepts are represented by means of typicality-based features, like in the $\pfl$ logic.}. 

Starting from a list of $135$ words that human subjects rated as proper \textit{emotion words}, Shaver and colleagues asked $100$ participants to gather these words by similarity, then computed clusters of words that represent subordinate emotions categories. These clusters, detected based on co-occurrence relations, are characterized by heterogeneity and by the intrinsic fuzziness of the terms they contain: for example, one cluster contained, among others, related words such ``arousal", ``desire" and ``lust". 
A higher level of clustering, roughly corresponding to the so called basic or primary emotions in the literature, was then detected: \textit{love}, \textit{joy}, \textit{surprise}, \textit{anger}, \textit{sadness}, and \textit{fear} were the resulting basic prototypical categories. The $6$ basic emotion prototypes can be further identifiable as positive or negative emotions at a higher level and each prototype emotion subsumes a list of emotion words, obtained by merging the subordinate clusters they subsume. 

In this  experiment, we have used the Shaver model of emotion as a lexical base for generating the prototypical emotional concepts with $\pfl$ (thus substituting, de facto, the NRC lexicon used in the previous experiment). 
As mentioned above, the Shaver model of emotion is particularly compliant with the overall assumptions of $\pfl$ since it provides the typical words associated to emotional concepts, collected from empirical data. It, however, differs from Pluthick's model, since it does not provide any compositional procedure to generate compound emotions by design. As a consequence of this state of affairs, in order to test the generative capacity of $\pfl$ also in this setting, we re-used Plutchik's model for determining the compositional rules of emotion combination. However, this time, the lexicon used  was provided by the Shaver model (and not obtained from the NRC). 

Plutchik's and Shaver's emotional models present some differences. In particular: Shaver's model considers only $6$ emotional concepts (of which  $5$ are also considered as primary emotions in the Pluthcik model). As a consequence of this state of affairs, we employed the $\pfl$ mechanisms only on such 5 emotional concepts (anger, fear, joy, sadness and surprise). 
As the Shaver model provides many lexical terms for each emotion, we selected a subset of terms. To be consistent to what we did with the NRC Lexicon, we selected a maximum of $6$ terms as typical properties. The proababilistic rating of the words selected form the Shaver's model were obtained by the intensity ratings already provided in the NRC Lexicon. As a result, the typical properties selected to be part of an emotion prototype were the Shaver terms having a higher intensity in NRC.

In Table \ref{shaver} are the results of the different reclassifications.
For the compound emotions generated starting from both Shaver's lexicon and the NRC lexicon, it emerges that Shaver's lexicon performs better on the ArsMeteo dataset, which is the most diversified one. For $4$ out of $10$ reclassifications with compound emotions, it also obtains better results on the RaiPlay dataset, while only in $3$ cases obtains more reclassifications than the NRC dataset. This datum overall shows that there are minimal differences in the adoption of the different lexicons using the provided pipeline in terms of number of reclassifications and that, overall, the combinatorial mechanisms used by DEGARI for emotion reclassification are, for these two lexicons, relatively independent from the particular lexicon used.

\begin{table*}
\caption{Reclassification results using Shaver lexicon, compared to the original NRC lexicon}
\vspace{-0.2cm}
\label{shaver}
\footnotesize
\centering
\begin{tabular}{l|r r|r r|r r}
\toprule
\rule[-1ex]{0pt}{2.2ex}  & \multicolumn{2}{p{2.2cm}|}{\textbf{ArsMeteo\newline(9171 artworks)}}  & \multicolumn{2}{p{2.2cm}|}{\textbf{RaiPlay\newline(4612 media items)}} & \multicolumn{2}{p{2.2cm}}{\textbf{WikiArt Emotions\newline(4105 artworks)}}  \\ 
\midrule
\rule[-1ex]{0pt}{2.2ex} \textbf{Emotion} & \textbf{NRC} & \textbf{Shaver} & \textbf{NRC} & \textbf{Shaver} & \textbf{NRC} & \textbf{Shaver} \\ 
\midrule
\rule[-1ex]{0pt}{2.5ex} {\textbf{awe (fear\_surprise)}} & $63$ & $117$ & $132$ & $7$ & $508$ & $0$  \\ 
\midrule
\rule[-1ex]{0pt}{2.5ex} {\textbf{delight (joy\_surprise)}} & $52$ & $71$ & $117$ & $2$ & $1672$ & $0$  \\
\midrule
\rule[-1ex]{0pt}{2.5ex} {\textbf{delight (surprise\_joy)}} & $52$ & $57$ & $117$ & $88$ & $1672$ & $0$  \\
\midrule
\rule[-1ex]{0pt}{2.5ex} {\textbf{despair (fear\_sadness)}} & $19$ & $49$ & $20$ & $132$ & $0$ & $508$  \\
\midrule
\rule[-1ex]{0pt}{2.5ex} {\textbf{disapproval (sadness\_surprise)}} & $58$ & $83$ & $64$ & $2$ & $404$ & $0$  \\
\midrule
\rule[-1ex]{0pt}{2.5ex} {\textbf{envy (anger\_sadness)}} & $25$ & $30$ & $21$ & $49$ & $0$ & $508$  \\
\midrule
\rule[-1ex]{0pt}{2.5ex} {\textbf{guilt (fear\_joy)}} & $22$ & $44$ & $73$ & $0$ & $1618$ & $0$  \\
\midrule
\rule[-1ex]{0pt}{2.5ex} {\textbf{guilt (joy\_fear)}} & $24$ & $45$ & $73$ & $117$ & $1618$ & $1672$  \\
\midrule
\rule[-1ex]{0pt}{2.5ex} {\textbf{outrage (anger\_surprise)}} & $41$ & $49$ & $46$ & $20$ & $508$ & $0$  \\
\midrule
\rule[-1ex]{0pt}{2.5ex} {\textbf{pride (anger\_joy)}} & $20$ & $27$ & $72$ & $64$ & $1618$ & $404$  \\
\midrule
\rule[-1ex]{0pt}{2.5ex} {\textbf{TOTAL}} & $324$ & $572$ & $618$ & $481$ & $7946$ & $3092$  \\
\bottomrule
\end{tabular}
\vspace{7pt}
\end{table*}

\subsection{User Study}

The goal of the user study was to assess the acceptance of the emotion categories suggested by DEGARI, with the ultimate goal of using the reclassifications produced by the system to improve the annotation of artworks and media, and, consequently, the applications which depend on it, such as personalization and recommendation.

\smallskip

\noindent \textit{Methods and material}. The user study consisted in an online questionnaire (in Italian). The questionnaire contained $10$ items, each accompanied by an image, or, for the multimedia items, by the film poster, accompanied by the link to the online player for watching the content. For each item, the users received two questions: the first question (Question $1$) asked them to rate the association of the item with the emotional category provided by DEGARI on 10-point scale; the second question (Question $2$)  asked them to associate the item to additional emotion categories,  taken from the Plutchik model. 
Users were divided in $3$ groups, each corresponding to a different set of $10$ items. For each dataset, the selected items were the ones ranked higher by DEGARI for each generated compound emotional category.
%

\smallskip 

\noindent \textit{Participants and procedure}. The study involved $44$ users ($23$ females, $21$ males). Concerning the age groups, $2$ users were below $18$; $17$ users were in the  $19$-$35$ age range; $12$ in the  $36$-$50$ range; $10$ in the  $51$-$70$ range; $3$ were older than $70$.
Users were randomly assigned to the questionnaires. The first questionnaire was filled out by $12$ users; the second questionnaire was filled out by $20$ users; the third questionnaire was assigned to $12$ users. As a result, $440$ ratings and $1065$ emotion categories were collected.

\smallskip

\noindent \textit{Results and analysis}. Concerning  Question $1$, the average rating assigned by the users to the emotion category proposed by Degari was $6.3$, with only slight differences between the datasets (see Table \ref{t:ratings}). The average rating was $6.47$ for ArsMeteo, $6.42$ for RaiPlay, and $6.12$ for WikiArt. 
The standard deviation was $2.41$ ($2.48$ for ArsMeteo, $2.47$ for WikiArt, and $2.1$ for RaiPlay), suggesting that the differences in ratings were limited.
Also, the median rating is $7$ for all data sets, with only $2$ proposed emotion categories ($1$ from Arsmeteo and $1$ from WikiArt) rated below $5$.


\begin{table*}
\caption{User ratings of the emotions proposed by DEGARI}
\label{t:ratings}
\centering
\begin{tabular}{c|c|c|c|c}
\toprule
\rule[-1ex]{0pt}{2.5ex}  & {\textbf{ArsMeteo}}  & {\textbf{WikiArt}} & {\textbf{RaiPlay}}  & {\textbf{All}}  \\ 
\midrule
\rule[-1ex]{0pt}{2.5ex} {\textbf{Average rating}} & $6.47$ & $6.12$ & $6.42$ & $6.32$ \\ 
\midrule
\rule[-1ex]{0pt}{2.5ex} {\textbf{Standard deviation}} & $2.48$ & $2.47$ & $2.1$ & $2.41$ \\ 
\midrule
\rule[-1ex]{0pt}{2.5ex} {\textbf{Median}} & $7$ & $7$ & $7$ & $7$ \\ 
\bottomrule
\end{tabular} 
\vspace{7pt}
\end{table*}

\noindent Concerning Question $2$ (namely, the additional emotions attached by the users to the items),
$308$ were attached to the items in ArsMeteo,	$308$ to the items in WikiArt, and $449$ to the items in RaiPlay, yielding $1065$ user emotion categories. The average number of emotion categories per users was $24.2$. In order to investigate the overlapping between the set of emotion categories proposed by DEGARI for each item (apart for the top ranked category tested through Question $1$), we compared the emotion categories selected by the users with the ones proposed by DEGARI (Table \ref{t:tags}). Data show that 22.05\% of the emotion categories additionally proposed by DEGARI for each item matched those selected by the users, with a higher value for ArsMeteo  ($27.78$\%), and a lower value for WikiArt ($20$\%) and RaiPlay ($19.51$\%). This datum is a positive one since it concerns the non-top ranked emotional categories suggested by the system (for which the degree of acceptability by the users was always above $5$ out of a $10$-point scale except for $2$ items, and with a median of $7$ for every considered dataset). 

\begin{table*}
\caption{Overlapping of user tags with DEGARI emotions}
\label{t:tags}
\centering
\begin{tabular}{c|c|c|c|c}
\toprule
\rule[-1ex]{0pt}{2.5ex}  & {\textbf{ArsMeteo}}  & {\textbf{WikiArt}} & {\textbf{RaiPlay}}  & {\textbf{All}}  \\ 
\midrule
\rule[-1ex]{0pt}{2.5ex} {\textbf{User tags}} & $308$ & $308$ & $449$ & $1065$ \\ 
\midrule
\rule[-1ex]{0pt}{2.5ex} {\textbf{Proposed emotions}} & $36$ & $50$ & $41$ & $127$ \\ 
\midrule
%
\rule[-1ex]{0pt}{2.5ex} {\textbf{Overlapping}} & $27.28$\% & $20$\% & $19.50$\% & $22.05$\% \\ 
\bottomrule
\end{tabular} 
\vspace{7pt}
\end{table*}

To conclude, the collected data suggest that the emotion categories proposed by DEGARI  as a result of the reclassification process are generally accepted by the users, with few exceptions that deserve further investigation. 
The acceptance is clear for the top ranked emotion, but 
a satisfactory
degree of acceptance can be inferred  also for the remaining suggested categories, for which an overlapping of $20$\% and more with the user tags has been found in all datasets.

\section{Discussion and Conclusion}\label{final}

In this paper, we presented DEGARI: an explainable AI system relying on the $\pfl$ Description Logics and on the ArsEmotica knowledge base to generate, according to Plutchik's theory of emotion, compound emotional concepts starting from the basic ones. Such newly created categories, characterized by lexicon-based typical features, are then used in DEGARI to reclassify, in an emotional settings, the items of three different datasets.
The novelty of this system relies on the fact that DEGARI is, to the best of our knowledge, the first emotion-oriented system employing a white box approach to emotion classification based on the  human-like conceptual combination framework proposed in the $\pfl$ logic. The explainability requirement comes for free as a consequence of this logic-based approach, as shown in Figure \ref{explanation}.

Overall, the white box approach proposed by DEGARI for emotionally-driven content reclassification could be useful for addressing the very well known filter bubble effect \cite{bubble} in recommender systems, by introducing seeds of serendipity in content discovery by users. 
One fundamental discussion about the applicability of DEGARI in practice is whether or not it represents a truly innovative technical solution for an emotion-based recommender system. According to \citeauthor{Sohaili2017}  \cite{Sohaili2017}, recommender systems ``try to identify the need and preferences of users, filter the huge collection of data accordingly and present the best suited option before the users by using some well-defined mechanism''. Despite the huge amount of proposals, the main families of recommender systems can be identified as based on: i) collaborative filtering; ii) content-based filtering; iii) hybrid filtering. At their core of functioning, collaborative filtering exploits similarities of usage patterns among mutually affine users, while content-based filtering exploits content similarity. DEGARI by definition falls into the latter category since in its current form it uses content description (obtained in different ways) as the input. From the technical point of view, however, it differs from the current mainstream approaches that are mostly based on 
the comparison and matching of visual and perceptual features of the content \cite{bdcc2018,Deldjoo2018}. In practice, our approach adds a logic layer capable of mapping and representing - in a commonsense and cognitively compliant fashion \cite{lieto2021cognitive} - new emotional categories which can be used to affect user preferences and content consumption in a way that cannot be derived from the pure statistical analysis of content and/or the comparison of similar users.
Moreover, the proposed approach has been applied to a well-known model, the Plutchik circumplex model of emotions \cite{plutchik1980general} and to two different emotional lexica (NRC and Shaver's), but could in principle be applied to other models which organize emotions by similarity, opposition and composition, such as for example the extended version of the Hourglass model used in SenticNet \cite{susanto2020hourglass}. Being independent from the specific application model and type of expression, this approach can work effectively in different domains, as shown by its use on the datasets of artworks and media illustrated in this paper. In this sense, it can promote the interoperability of affective annotations and the cross-domain reuse of techniques and methods.

In the future work, we plan to extend the evaluation currently conducted in the form of a user study to a large scale one to further validate the effectiveness of the proposed approach. We also plan to extend the applications of this system to different domains. A first extension will be in the field of the emotional-oriented recommendation of artworks within Museums and cultural heritage sites (this is a work currently under development within the H2020 European SPICE project\footnote{\url{https://spice-h2020.eu/}}). In addition, also the field of music recommendation represents a current area of investigation.

From a technical perspective, in future research, we aim at studying the application of optimization techniques in \cite{cplint} in order to improve the efficiency of the DEGARI knowledge generation system. Secondly, we aim at considering more accurate and multimodal descriptions of artistic and media items, by exploiting Automatic Speech Recognition data and semantic visual categories extracted from video and audio channels of the content.
Finally, as mentioned, we plan to improve the provided recommendations by justifying the content reclassification (and the derived recommendations) based on the probabilistic ranks assigned to the shared features between the generated emotion and the items being reclassified.

\section*{Acknowledgements}

This work has been partially funded from the European Union’s Horizon 2020 research and innovation program under grant agreement
No 870811 (SPICE Project).

\bibliographystyle{apalike}


\bibliography{mybibfile}

%
%


\end{document}